\documentclass[unnumsec,webpdf,modern,large]{oup-authoring-template}
\onecolumn % for one column layouts

\graphicspath{{Figures/}}

\usepackage[right]{lineno}

\theoremstyle{thmstyleone}%
\theoremstyle{thmstyletwo}%
\theoremstyle{thmstylethree}%

\begin{document}

\journaltitle{}
\DOI{}
\copyrightyear{}
\pubyear{}
\access{}
\appnotes{}

\firstpage{1}

%\subtitle{Subject Section}

\title[AI Agents for Auto Science Discovery]{Aleks: AI powered Multi Agent System for Autonomous Scientific Discovery via Data-Driven Approaches in Plant Science}

\author[1]{Daoyuan Jin}
\author[2]{Nick Gunner}
\author[3]{Niko Carvajal Janke}
\author[3]{Shivranjani Baruah}
\author[3]{Kaitlin M. Gold}
\author[2,$\ast$]{Yu Jiang}

\authormark{}

\address[1]{\orgdiv{School of Electrical and Computer Engineering}, \orgname{Cornell University}, \orgaddress{\street{Ithaca}, \state{New York}, \postcode{14850}, \country{USA}}}
\address[2]{\orgdiv{Horticulture Section, School of Integrative Plant Science, Cornell AgriTech}, \orgname{Cornell University}, \orgaddress{\street{Geneva}, \state{New York}, \postcode{14456}, \country{USA}}}
\address[3]{\orgdiv{PPPMB Section, School of Integrative Plant Science, Cornell AgriTech}, \orgname{Cornell University}, \orgaddress{\street{Geneva}, \state{New York}, \postcode{14456}, \country{USA}}}
%\address[4]{\orgdiv{Department}, \orgname{Organization}, %\orgaddress{\street{Street}, \postcode{Postcode}, \state{State}, %\country{Country}}}

\corresp[$\ast$]{Corresponding author. \href{email:yujiang@cornell.edu}{yujiang@cornell.edu}}

%\received{Date}{0}{Year}
%\revised{Date}{0}{Year}
%\accepted{Date}{0}{Year}

%\editor{Associate Editor: Name}

%\abstract{
%\textbf{Motivation:} .\\
%\textbf{Results:} .\\
%\textbf{Availability:} .\\
%\textbf{Contact:} \href{name@email.com}{name@email.com}\\
%\textbf{Supplementary information:} Supplementary data are available at \textit{Journal Name}
%online.}

\abstract{Modern plant science increasingly relies on large, heterogeneous datasets, but challenges in experimental design, data preprocessing, and reproducibility hinder research throughput. Here we introduce Aleks, an AI-powered multi-agent system that integrates domain knowledge, data analysis, and machine learning within a structured framework to autonomously conduct data-driven scientific discovery. Once provided with a research question and dataset, Aleks iteratively formulated problems, explored alternative modeling strategies, and refined solutions across multiple cycles without human intervention. In a case study on grapevine red blotch disease, Aleks progressively identified biologically meaningful features and converged on interpretable models with robust performance. Ablation studies underscored the importance of domain knowledge and memory for coherent outcomes. This exploratory work highlights the promise of agentic AI as an autonomous collaborator for accelerating scientific discovery in plant sciences.}

% \boxedtext{
% \begin{itemize}
% \item Key boxed text here.
% \item Key boxed text here.
% \item Key boxed text here.
% \end{itemize}}

\maketitle

%\linenumbers

\section{Introduction}
Modern plant science research has undergone a significant shift toward data-driven approaches, which has fueled the rapid development of new sensing technologies and computational tools in fields such as computational biology and digital agriculture. However, this paradigm also reveals a critical bottleneck: the challenge of analyzing massive, heterogeneous datasets to extract meaningful findings and knowledge. Issues of reproducibility and long turn-around times remain major barriers, often forcing researchers to devote substantial effort to “leg work” such as experimental design and data preprocessing instead of focusing on intellectual challenges. These limitations ultimately hinder the throughput required to address key scientific questions with the depth and efficiency they demand.\\

Intelligent agent (IA) research focuses on developing systems that can autonomously perceive information, perform reasoning and decision-making, execute required actions, and continuously learn and adapt to changes in both the agent itself and its surrounding environment. Recent advances in foundation models such as large language models (LLMs) and retrieval augmented generation (RAG) have greatly expanded the potential of IA systems. These advances allow agents to better interpret environments through natural language understanding and long context reasoning, while also acquiring new knowledge from continuously updated information. As a result, IA systems built upon these new AI technologies, also known as Agentic AI, are now able to address complex challenges beyond automating routine tasks. This shift opens promising opportunities for advancing critical areas of human society, including scientific discovery.\\

In general, large language models (LLMs) are less accurate for questions that require specialized expertise and knowledge, and therefore cannot be directly applied to advanced tasks such as scientific discovery or data analytics. To address this limitation, researchers have proposed methods to specialize LLMs, such as fine-tuning with domain-specific content, so that they can achieve higher accuracy in answering domain-related questions. However, this approach requires substantial effort to obtain domain training datasets and also demands considerable computational and energy resources. An alternative is to integrate LLMs in intelligent agent (IA) systems with external memory systems such as retrieval augmented generation (RAG), vector databases, or knowledge graphs, which can enhance their capabilities. Still, a single agent is rarely able to cover a wide range of specialized expertise, and thus remains limited to a specific domain. Multi-agent systems (MAS) offer a promising solution in which multiple IAs, each specialized in a particular domain, can work collectively as a team to address complex problems.\\

Recent advances in AI-powered agent systems have demonstrated their potential to automate different aspects of scientific research across diverse domains. In data science, Yao et al. (2023) \cite{yao2023react} introduced ReAct, a prompting paradigm that integrates reasoning and acting in large language models (LLMs), enabling dynamic plan generation, fact grounding, and exception handling. This approach mitigated hallucinations common in chain-of-thought reasoning and achieved strong performance on multi-hop question answering, fact verification, and interactive environments, highlighting the benefits of combining internal reasoning with external interactions for general decision-making. Schmidgall et al. (2025) \cite{schmidgall2025agent} proposed Agent Laboratory, an open-source LLM-agent framework capable of autonomously executing the machine learning research workflow, including literature review, experiment design, execution, and report writing. While technically autonomous and competitive on benchmark challenges, its outputs often fell short of publishable quality and were better suited for rapid prototyping and research summaries. Similarly, Trirat et al. (2025) \cite{trirat2024automl} introduced AutoML-Agent, a multi-agent framework for automated machine learning pipelines encompassing preprocessing, model selection, hyperparameter optimization, and ensembling. Their results highlighted the strength of modular agent-based orchestration, though the system lacked domain-oriented guidance and interpretability. Swanson et al. (2025) \cite{swanson2025virtual} presented the Virtual Lab, a multi-agent framework comprising a Principal Investigator, domain-specific scientist agents, and a Scientific Critic to facilitate interdisciplinary research. In their SARS-CoV-2 nanobody design case study, the framework produced 92 candidates, two of which showed improved binding in wet-lab validation. However, human researchers remained responsible for agenda setting, tool selection, and experimental validation, underscoring that the system supports productive AI–human collaboration rather than fully autonomous discovery. Collectively, these efforts illustrate the growing capabilities of AI-powered agent systems, while also revealing persistent challenges in autonomy, domain specialization, and scientific rigor.\\

For plant sciences, an early effort is the PhenoAssistant system \cite{chen2025phenoassistant}, which employs a conversational agent design to support plant phenotyping. In this framework, a manager LLM coordinates phenotype extraction, analysis, and visualization through specialized vision models and automated pipelines. The system demonstrates strong performance in tool and model selection; however, it operates in a user-in-the-loop mode, where higher-level reasoning and task selection are still directed by humans. As a result, PhenoAssistant provides valuable automation for computational execution but does not yet achieve full autonomy as expected for AI-powered IA or MAS systems. Therefore, it remains open to understand if an AI-powered MAS systems could potentially achieve full autonomy for research questions in plant sciences.\\

The overarching goal of this study is to develop Aleks, an AI-powered MAS designed to autonomously address scientific discovery questions in plant sciences. The specific objectives are to i) design and implement Aleks in order to achieve full autonomy in data-driven plant science research, ii) evaluate its efficacy through a case study focused on plant disease research, and iii) provide insights into future directions of AI-powered MAS for advancing autonomous scientific discovery in plant sciences.\\

\section{Aleks System Design and Architecture}
\subsection{Overall System Concept and Implementation}
\begin{figure}[h!]
    \centering % Centers the image and caption
    \includegraphics[width=0.5\textwidth]{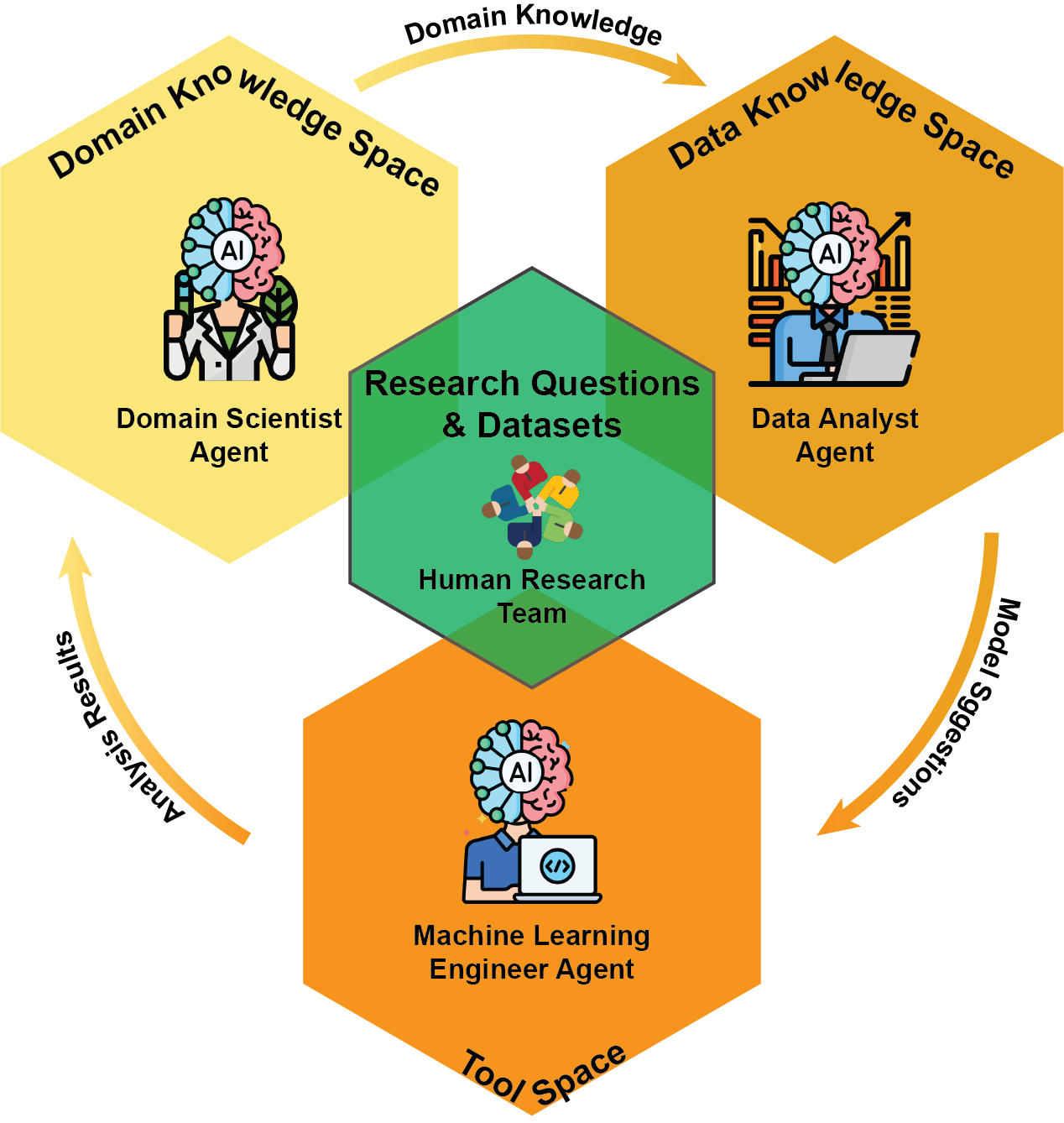}
    \caption{Conceptual framework of Aleks. Human researchers provide a scientific question and an associated dataset to Aleks. Within Aleks, three domain expertise are represented as autonomous agents: the domain scientist, the data analyst, and the machine learning engineer. Knowledge flows from domain understanding, to data analysis, and then to machine learning, with each stage translated and coordinated through interactions among the agents.}
    \label{fig:sysarch}
\end{figure}

We designed Aleks, an AI-based multi-agent system (MAS) that integrates domain expertise and data analytics to autonomously address scientific questions through data-driven approaches (Figure~\ref{fig:sysarch}). Conceptually, Aleks consists of multiple LLM-powered agents, each specialized in domain knowledge, data science, or machine learning engineering. These agents collaborate in a structured manner to translate domain expertise into modeling strategies and incorporate the resulting insights into machine learning implementations. Outcomes from machine learning experiments are then fed back to the domain agent, creating an iterative loop that continues until either a satisfactory solution is achieved or the allocated research budget is exhausted. Aleks accepts scientific questions posed in natural language by human researchers, together with the associated datasets, and employs its agents to generate answers derived from the data.\\

\begin{figure}[h!]
    \centering % Centers the image and caption
    \includegraphics[width=0.7\textwidth]{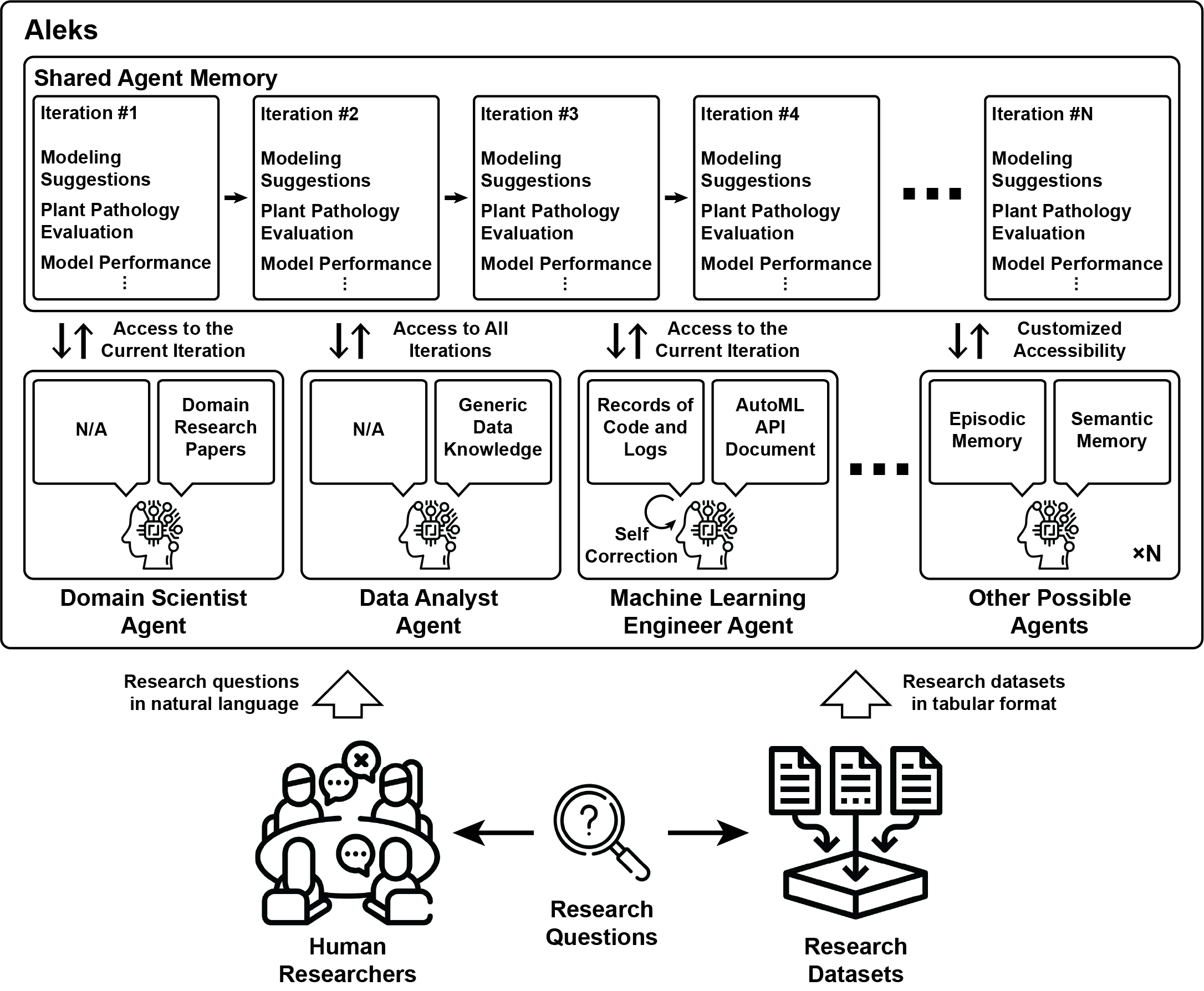}
    \caption{Aleks system design and implementation. In the present study, Aleks consists of three specialized agents: a plant pathologist, a data analyst, and a machine learning engineer. Each agent contributes distinct expertise and interacts through a shared memory that can be accessed either within the current iteration or across the entire history of interactions. Although only three agents are implemented here, the framework is designed to accommodate the addition of further specialized agents in the future.}
    \label{fig:syscomponents}
\end{figure}

To implement this concept, Aleks is structured around three specialized agents that communicate through a shared memory architecture (Figure~\ref{fig:syscomponents}). Each agent has distinct responsibilities and functions: the domain scientist (DS) agent provides domain-specific knowledge for a given scientific question and verify biology relevance of proposed analytical approaches and results; the data analyst (DA) agent generates modeling suggestions by considering knowledge and feedback from the domain scientist agent; and the machine learning engineer (MLE) agent implements machine learning modeling decisions from the DA agent and generates experimental results for review. A shared memory system is implemented to archive experimental records including task descriptions, datasets, modeling suggestions, results, and feedback. Depending on the communication needs, the agents are customized with the accessbility to this shared memory to enable communication among them. Additionally, each agent can have its own episodic and semantic memory to specialize its knowledge base or skills for specific tasks that would require iterative refinement by that agent itself like debugging codes. Through iterative processes, Aleks balances automated exploration with interpretability and domain relevance, providing a scalable framework for data-driven scientific research.\\

\subsection{Individual Intelligent Agents}
\subsubsection{Domain Scientist Agent}
To incorporate domain knowledge into the modeling process, we implemented a DS agent that evaluates the modeling suggestions and machine learning results by considering domain-specific knowledge. The DS agent maintains a structured knowledge base extracted from scientific literature provided by human researchers, using an LLM to summarize key domain variables and causal relationships and saving the resultant context into the semantic memory of the DS agent.\\

After each modeling iteration, the DS agent will be called with a summary of the experiment including selected features and the latest modeling report. The DS agent assesses whether the result is meaningful within the domain context and whether the dataset has been sufficiently utilized to answer the given scientific question. The DS agent is also encouraged to prioritize modeling strategies that use fewer (higher robustness) and more interpretable (higher domain knowledge relevance) features when possible, and to consider domain-specific biases such as the relative cost of false positives versus false negatives. Based on this context, the DS agent returns a concise critique in natural language description and suggests domain-informed improvements to data preprocessing and/or feature engineering. This feedback loop enables iterative, knowledge-guided model refinement while preserving the modular and interpretable structure of the agentic framework.

\subsubsection{Data Analyst Agent}
We developed a DA agent to autonomously refine analysis strategies on input datasets. Given the input scientific question and corresponding datasets, the DA agent systematically proposes modeling strategies, evaluates outcomes, and improves data preprocessing and feature engineering through multiple iterations. Feature engineering includes both new feature derivation and feature selection. \\

At each iteration, the DA agent constructs a prompt incorporating the dataset's metadata, sample rows, and full experiment history, including previously selected features, preprocessing notes, problem formulation (i.e., model type), and performance metrics. The agent is explicitly encouraged to explore possible problem formulation options during early iterations and determine the most appropriate framing based on empirical evidence. In the present study, the formulation options are classification and regression.\\

Each Aleks' process iteration concludes with a decision to either stop (if performance is satisfactory) or continue with new modeling suggestions. Once the agent is satisfied or reaches the research budget (the maximum iteration of Aleks processes), it generates a final summary report describing the recommended features, preprocessing steps, model type, justification on the recommendation, and analysis code for human researchers.

\subsubsection{Machine Learning Engineer Agent}
We developed an MLE agent to automate the development of computer programs for model training and evaluation on the given datasets. The MLE agent firstly reviews human researcher questions and modeling suggestions from the DA agent, and subsequently previews the dataset(s) by sampling records at a fixed interval to provide representative context. Based on this preview, the MLE agent constructs a constrained natural language prompt that instructs its LLM to generate executable Python code. Each script is saved with a timestamped filename in a dedicated subdirectory for version control and then executed in a subprocess that streams outputs in real time while capturing them for downstream use.\\

The key for the MLE agent is the success rate of generating executable codebase for machine learning training and evaluation. To this end, the MLE agent has its own semantic and episodic memory. Its semantic memory is established with the API document of possible libraries. Meanwhile, prompts are added to the semantic memory to enforces computer memory use limit (e.g., avoiding nested loops with $O(N^2)$ efficiency) and upper bound of program runtime and to prioritize vectorized or tree-based methods for spatial feature engineering. Its episodic memory is constructed with structure runtime messages from the execution of the generated code. If the script execution fails, error messages are fed back to the prompt to guide the MLE refine the code until a valid solution is obtained or retry limits are reached. This design provides an automated pipeline that connects dataset inspection, code generation, execution, and error handling within a controlled environment, ensuring both reproducibility and stability while executing autonomously.\\

\subsection{Shared Memory for Cross Agent Communication}
A central memory system is employed to sequentially store structured records of the entire process, from domain knowledge incorporation to machine learning outputs, and to share these records across agents according to their access privileges. These records include the initial dataset and task description provided by human researchers, along with all subsequent iterative processes carried out by Aleks. Each iteration is stored with four key elements: experiment iteration index, modeling suggestions, machine learning results, and domain-specific feedback to the suggestions and results.\\

To mimic effective interdisciplinary collaboration, agents interact with the central memory system in distinct ways. The domain scientist agent accesses the scientific question and corresponding datasets from human researchers and current modeling results. Grounded in both data and outcomes, this agent provides scientifically meaningful interpretations that guide the iterative process of Aleks. The data analyst agent has access to the complete experimental history, enabling it to reason across iterations, identify patterns, and propose refined adjustments for subsequent modeling cycles by incorporating both domain knowledge from the domain agent and general data science knowledge from itself. The machine learning engineer agent interacts with a working memory view limited to the dataset, task description, and modeling suggestions of the current iteration. This scoped access ensures the agent focuses on executing and optimizing the assigned modeling task without distraction from irrelevant historical details.\\

Through this design, the memory system functions as a central communication hub to maintain continuity across iterations, prevent information loss, and deliver tailored contextual views to each agent. By combining full-history access with selective memory exposure, this memory system design balances efficiency with interpretability in cross-agent communication.\\

\section{Plant Tissue Sampling for Disease Monitoring}
\subsection{Background in Grape Red Blotch Disease}
Grapevine red blotch disease (GRBD), caused by grapevine red blotch virus (GRBV), threatens wine‐grape yield and quality by disrupting carbon allocation, delaying ripening, and depressing sugar accumulation~\cite{sudarshana2015grapevine}. Left unmanaged, its lifetime economic burden can reach ~\$68,548 per hectare over a 25-year vineyard lifespan, underscoring the imperative for earlier, more systematic detection in high-value regions such as Napa Valley~\cite{cieniewicz2025grapevine}.\\

GRBV symptoms are difficult to diagnose reliably because they usually appear late in the season, when management options are limited, and are easily confused with nutrient stress, other pathogens, or abiotic injury \cite{cieniewicz2025grapevine}. Symptom expression is uneven within vines, with only a subset of shoots or leaves affected at any given time, making scouting prone to false negatives and misclassification~\cite{laroche2025grapevine}. These diagnostic challenges are further compounded by the epidemiology of GRBD, which involves progressive, spatially structured spread shaped by local aggregation, background inoculum, and vector movement by the three-cornered alfalfa hopper. This spatial heterogeneity means that symptom expression varies not only within vines but also across vineyard blocks, complicating surveillance and reinforcing the need for risk-based approaches to detection and management \cite{cieniewicz2017spatiotemporal, flasco2023distinct, flasco2023three, flasco2024investigating, flasco2025decade, jeger2023emerging}. Molecular assays (e.g., PCR/qPCR) remain the diagnostic gold standard, but they are costly and time-consuming, and their accuracy hinges on disciplined sampling protocols that account for within-vine heterogeneity—i.e., selecting the right tissue, at the right time, from the right place on the plant \cite{cieniewicz2025grapevine}.\\

Remote sensing and epidemiological modeling together provide a powerful basis for optimizing GRBD sampling strategies. Remote sensing enables the generation of risk maps at operational scales by capturing disease-induced perturbations in canopy biochemical and biophysical traits that alter optical signatures across the visible to shortwave infrared spectrum \cite{gold2021plant, jeger2024impact, mikaberidze2025cost}. Hyperspectral imaging can screen entire vineyard blocks, efficiently flagging candidate vines for confirmatory testing, and recent advances are extending detection from leaf-level proofs of concept to canopy-level applications under field condition \cite{sawyer2023phenotyping, laroche2025grapevine}. Airborne imaging spectroscopy has further demonstrated the capacity for scalable, presymptomatic detection across multiple pathosystems, streaming near real-time data layers for decision support \cite{zarcoprevisual, zarco2021divergent, galvan2023scalable, rubambiza2023toward}. Coupling these sensing tools with epidemiological modeling closes the loop between identifying where disease is likely present and directing where to sample next. \\

A crucial question to human plant pathologists for GRBD is how to design the best sampling strategy especially in terms of locations to maximize the molecular tests outcome that can be used to inform successive research efforts on disease epidemiological development, disease infection mechanisms, and possible management practices. In this case study, a multi-year vineyard dataset was compiled at a 10 m × 10 m grid level. For each grid cell, the annual number of symptomatic vines was recorded along with complementary features such as historical vegetation indices derived from remote sensing and partial information on vector presence and activity. We aimed to challenge Aleks to develop a machine learning solution that can use observations available for this problem to predict GRBD infection status and inform plant tissue sampling strategy to human researchers. It is noteworthy that the only input from human researchers was a single question description: “The goal is to predict redvine disease in 2023 or 2024.” alongwith the compiled dataset. All subsequent decisions regarding task framing, problem formulation (i.e., how to translate the question to a machine learning problem), feature engineering, and modeling were entirely and autonomously managed and conducted by Aleks.

\subsection{Aleks Configuration}
Aleks was configured to address the specific requirements of this case study. The DS agent was specialized as a plant pathologist by establishing its semantic memory with summaries of ten research papers selected by human researchers. This ensured that knowledge of GRBD and its epidemiological characteristics could directly inform subsequent data processing by the DA and MLE agents. The MLE agent was configured to exclusively use the auto-sklearn library for machine learning tasks. This choice was guided by two considerations: first, the GRBD records and remote sensing features were provided in tabular format well suited to auto-sklearn; second, the library is well scoped and thoroughly documented, which improved the agent’s success rate in automating coding tasks and allowed it to focus on its role within the multi-agent system rather than becoming hindered by coding or debugging. All other configuration parameters were maintained at their default settings. In particular, the LLM used in this study was DeepSeek Chat. 

\subsection{Deploying Aleks for Tissue Sampling Strategy Recommendation}
We designed four experiments to systematically evaluate Aleks in addressing the plant sampling strategy problem for GRBD (Table~\ref{tab:exp_config}). The first experiment assessed the reproducibility and robustness of Aleks in converging on solutions to research questions posed by human researchers. Since each agent in Aleks is powered by an LLM, which introduces variability in responses to identical prompts, it was essential to evaluate the impact of this stochasticity on system performance. For each research question, five independent runs were conducted using the same input prompts and corresponding datasets. This experiment also served as the baseline for subsequent ablation comparisons with alternative configurations of Aleks. \\

The second experiment evaluated the importance of incorporating domain-specific knowledge into data-driven research by testing a modified version of Aleks that included only the data analyst (DA) and machine learning engineer (MLE) agents, omitting the domain scientist (DS) agent. The third experiment examined the role of the shared memory system in cross-agent communication. In this configuration, Aleks was limited to sharing only the most recent iteration of experimental records rather than the complete history, thereby testing the contribution of long-term memory to coherent reasoning and decision-making across iterations. The final experiment explored a simplified global history design through a leaderboard mechanism that retained partially compressed reasoning traces alongside the full log. This experiment was intended to simulate scenarios in which extremely long experimental histories could exceed the context length of an LLM, while assessing whether a hybrid long–short history structure could sustain system performance.\\

For all experimental runs, Aleks was provided with identical research question and the corresponding dataset. The questions were framed as either “predict GRBD-infected grapevines in 2023” or “predict GRBD-infected grapevines in 2024.” The dataset itself contained information from both years and was input into Aleks without any preprocessing, such as splitting by year or filtering missing values. These preprocessing decisions were left entirely to Aleks as part of its reasoning and action process. The research budget, defined as a maximum of 20 iterations, was set based on results from a small-scale preliminary test. \\

\begin{table}[h!]
\centering
\begin{tabular}{cllcc}
\hline
Exp \# & Agents & Memory & Leaderboard & Sets \\
\hline
1 & DS, DA, and MLE agents & Full history & No  & 5 \\
2 & DA and MLE agents & Full history & No  & 1 \\
3 & DS, DA, and MLE agents & Single iteration history & No  & 1 \\
4 & DS, DA, and MLE agents & Full history & Yes & 1 \\
\hline
\end{tabular}
\caption{Experiment design for evaluating Aleks in plant sampling strategies for GRBD infection.}
\label{tab:exp_config}
\end{table}

\subsection{Evaluation Methods}
Unlike conventional machine learning experiments where evaluation metrics are predetermined, in our framework the choice of metrics was itself delegated to the Aleks system. After interpreting the task and defining the modeling problem, Aleks autonomously selected appropriate evaluation criteria and reported results accordingly. In most cases, the system opted for F1-score when formulating the task as a classification problem, and for coefficient of determination ($R^2$) when formulating the task as a regression problem. This design reflects the system’s capacity to reason about problem framing and determine suitable evaluation strategies without human specification. All reported metrics were manually verified for consistency with the experiment logs: accuracy, weighted F1, and confusion matrices for classification; and R² and RMSE for regression.\\

To address the interests of human scientists, we conducted cross-year performance testing on the best models generated by Aleks. Specifically, the optimal feature set identified by Aleks for the 2023 subset was used to train a model for predicting GRBD infection in 2024, and vice versa. In this way, models were evaluated not only on the year for which they were originally developed but also across years, providing an assessment of their generalizability. \\

\section{Results}
\subsection{Overall Performance of Aleks}

Overall, Aleks achieved the full autonomy of answering the questions from human researchers without any human interventions through repetitive runs. Once Aleks was given the question and corresponding dataset, it approached the question, formulated the question into a machine learning task (e.g., classification or regression), performed the analysis, and summarized results and proposed improvements for the next iteration autonomously.\\

\begin{figure}[ht!]
    \centering % Centers the image and caption
    \includegraphics[width=0.5\textwidth]{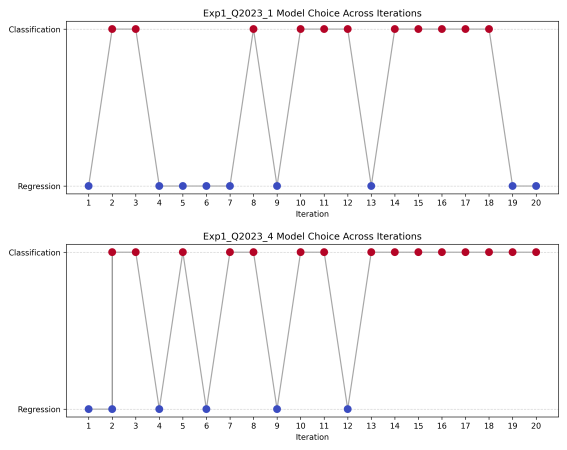}
    \caption{Modeling choice change plot for Exp1, 2023 prediction, repetition 1 and 4.}
    \label{fig5}  
\end{figure}

Aleks reliably converged on successful solutions for both the 2023 and 2024 prediction questions across repeated runs, despite differences in problem formulation strategies (Table~\ref{tab:model_performance}). With respect to problem formulation, Aleks demonstrated a strong and consistent preference for treating the 2024 prediction question as a regression task, regressing GRBD incidence on canopy and historical canopy traits (e.g., canopy area and enhanced vegetation index). In contrast, for the 2023 question, Aleks alternated between regression and classification formulations. Notably, within a single run Aleks often evaluated both formulations, sometimes even within the same iteration, before selecting the most promising approach (Figure~\ref{fig5}). This behavior suggests that Aleks engaged in active exploration of problem representations rather than converging prematurely on a single option, which helps mitigate potential biases introduced during LLM pretraining.\\

\begin{table}[h!]
\centering
\scriptsize
\begin{tabular}{l l l l l l l l l}
\hline
Experiment & Best Model Iteration & Total Runtime & Model Type & Accuracy & Weighted F1 & Confusion Matrix & R-square & RMSE \\
\hline
Exp1 2023 (1) & 8  & 129 min & Classification & 0.93 & 0.93 & [[2286, 66], [139, 447]] &  &  \\
Exp1 2023 (2) & 20 & 135 min & Regression     &      &      &                          & 0.6884 & 1.0272 \\
Exp1 2023 (3) & 14 & 79 min & Classification & 0.95 & 0.95 & [[3851, 80], [131, 462]] &        &        \\
Exp1 2023 (4) & 19 & 125 min & Classification & 0.95 & 0.95 & [[3867, 64], [140, 453]] &        &        \\
Exp1 2023 (5) & 20 & 119 min & Regression     &      &      &                          & 0.7475 & 0.9246 \\
Exp1 2024 (1) & 20 & 126 min & Regression     &      &      &                          & 0.6909 & 0.7294 \\
Exp1 2024 (2) & 19 & 114 min & Regression     &      &      &                          & 0.7726 & 0.6256 \\
Exp1 2024 (3) & 15 & 114 min & Regression     &      &      &                          & 0.7790 & 0.6168 \\
Exp1 2024 (4) & 17 & 124 min & Regression     &      &      &                          & 0.7574 & 0.6462 \\
Exp1 2024 (5) & 19 & 131 min & Regression     &      &      &                          & 0.8196 & 0.5572 \\
\hline
\end{tabular}
\caption{Summary of formulation options and modeling performance for GRBD infection prediction conducted by Aleks.}
\label{tab:model_performance}
\end{table}

Furthermore, Aleks consistently identified the correct label and appropriate features corresponding to the human scientist’s inquiry. For instance, when tasked with predicting 2023 GRBD-infected grapevines, Aleks correctly selected the column representing 2023 GRBD counts as the label and incorporated relevant predictors such as GRBD incidence and canopy traits from prior years. This ensured that no data leakage occurred and that the subsequent analyses remained valid.\\

\begin{figure}[h!]
    \centering % Centers the image and caption
    \includegraphics[width=0.5\textwidth]{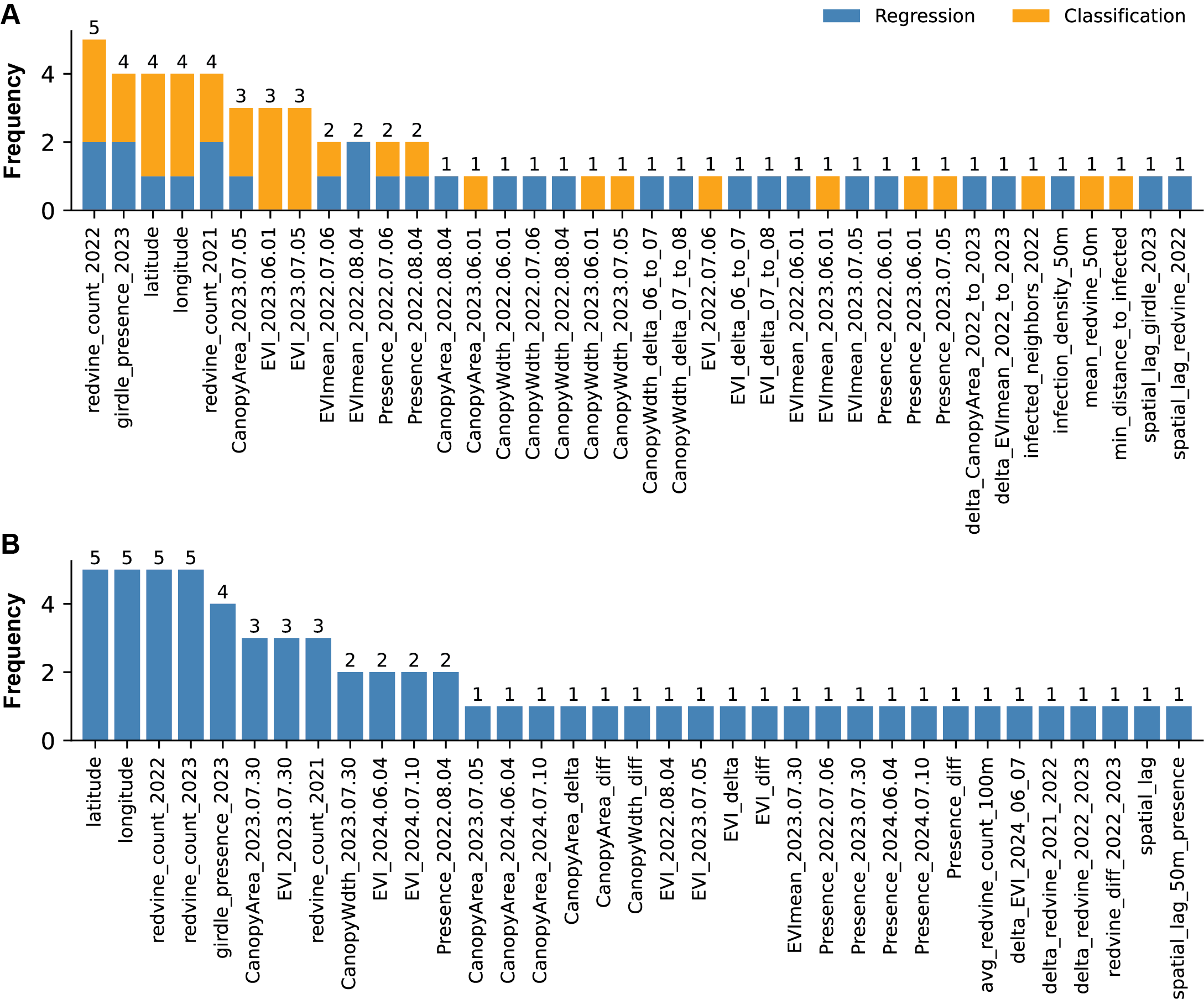}
    \caption{Feature frequency of the final selected models from five repeated experiments for each year’s prediction in Exp1.}
    \label{fig3}
\end{figure}

Aleks also steadily identified key features from the original input and attempted to propose new features across these repetitive runs for both questions (Figure~\ref{fig3}). Frequently used (frequency$\ge$ 60\%) original features included the counts of GRBD infected grapevines from prior years, geo-spatial coordinates, and canopy traits such as width, area, and vegetation indices (EVI). These features have been proven to be important not only to the performance of a machine learning task for GRBD infection prediction but also relevant to both the GRBD biology and epidermiology that have been reported in previous publications. This feature selection behavior demonstrated the efficacy of Aleks to incorporate complex domain knowledge to select relevant features for data-driven approaches to scientific questions. \\

\begin{figure}[ht!]
    \centering % Centers the image and caption
    \includegraphics[width=0.5\textwidth]{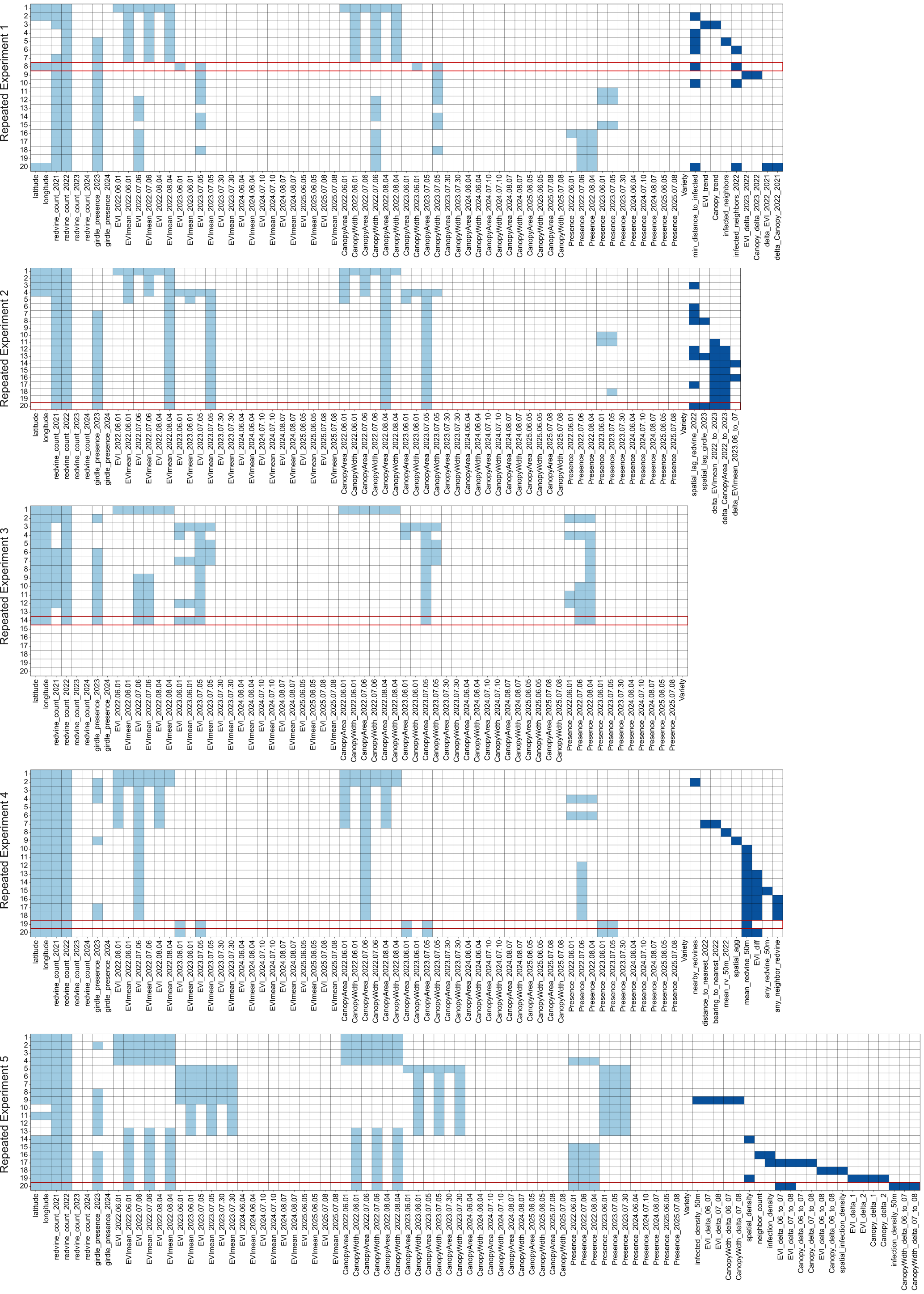}
    \caption{Feature selection map for five repeated runs in Exp1 for the 2023 prediction question. The heatmap illustrates the evolution of feature selection by Aleks across iterations. Light blue indicates raw features from the dataset (ordered consistently across all heatmaps), while dark blue represents derived features, arranged in the order in which they appeared during the experiment. }
    \label{fig4_1_1}
\end{figure}

\begin{figure}[ht!]
    \centering % Centers the image and caption
    \includegraphics[width=0.5\textwidth]{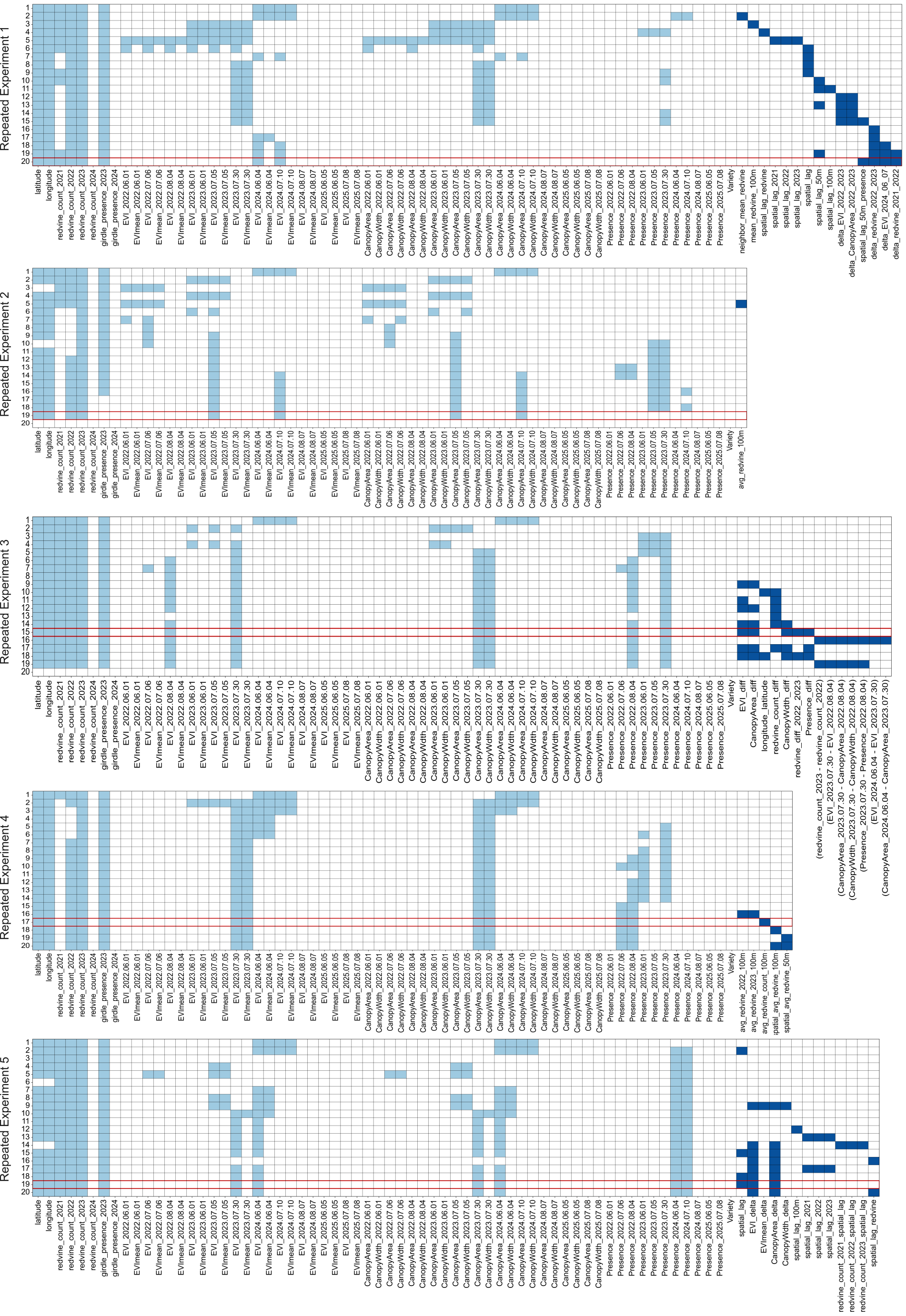}
    \caption{Feature selection map for five repeated runs in Exp1 for the 2024 prediction question. The heatmap illustrates the evolution of feature selection by Aleks across iterations. Light blue indicates raw features from the dataset (ordered consistently across all heatmaps), while dark blue represents derived features, arranged in the order in which they appeared during the experiment.}
    \label{fig4_1_2}
\end{figure}

In the early stages of its workflow, Aleks tended to rely on original features to explore different modeling strategies, often focusing on direct inputs such as GRBD infected grapevine counts or canopy traits (Figure~\ref{fig4_1_1} and Figure~\ref{fig4_1_2}). As iterations progressed, it gradually began to propose new traits that were derived from the original features and embedded with more domain knowledge. For example, instead of directly using geospatial coordinates, Aleks introduced the concept of GRBD infection lag, which incorporated a large spatial buffer zone to account for neighboring infected grapevines. These derived features reflected a deeper integration of plant science knowledge beyond pure data science operations. However, not all proposed features could be adopted. One challenge was that some suggested traits were drawn from research literature, such as hyperspectral images or vegetation indices, which were not available in the dataset. Another limitation was that certain suggestions were too abstract or loosely defined, making them difficult for the machine learning engineer agent to interpret and implement. As a result, only the practically available and clearly defined features were integrated into the subsequent data analysis iterations and considered by Aleks.\\

Aleks concluded each experiment run when it either found a satisfactory solution or reached the research budget limit (i.e., the maximum number of iterations), and subsequently produced an experiment report. The reported agent-selected model reflected Aleks' own recommendation by considering both data analysis performance and domain-specific knowledge, not necessarily the best performing model, but often one balancing predictive performance with domain interpretability. For example, models with fewer features were preferred to reduce overfitting risks. 

\subsection{Ablation Comparisons}
\subsubsection{Significance of the Domain Scientist Agent}
\begin{figure}[htbp]
    \centering % Centers the image and caption
    \includegraphics[width=0.5\textwidth]{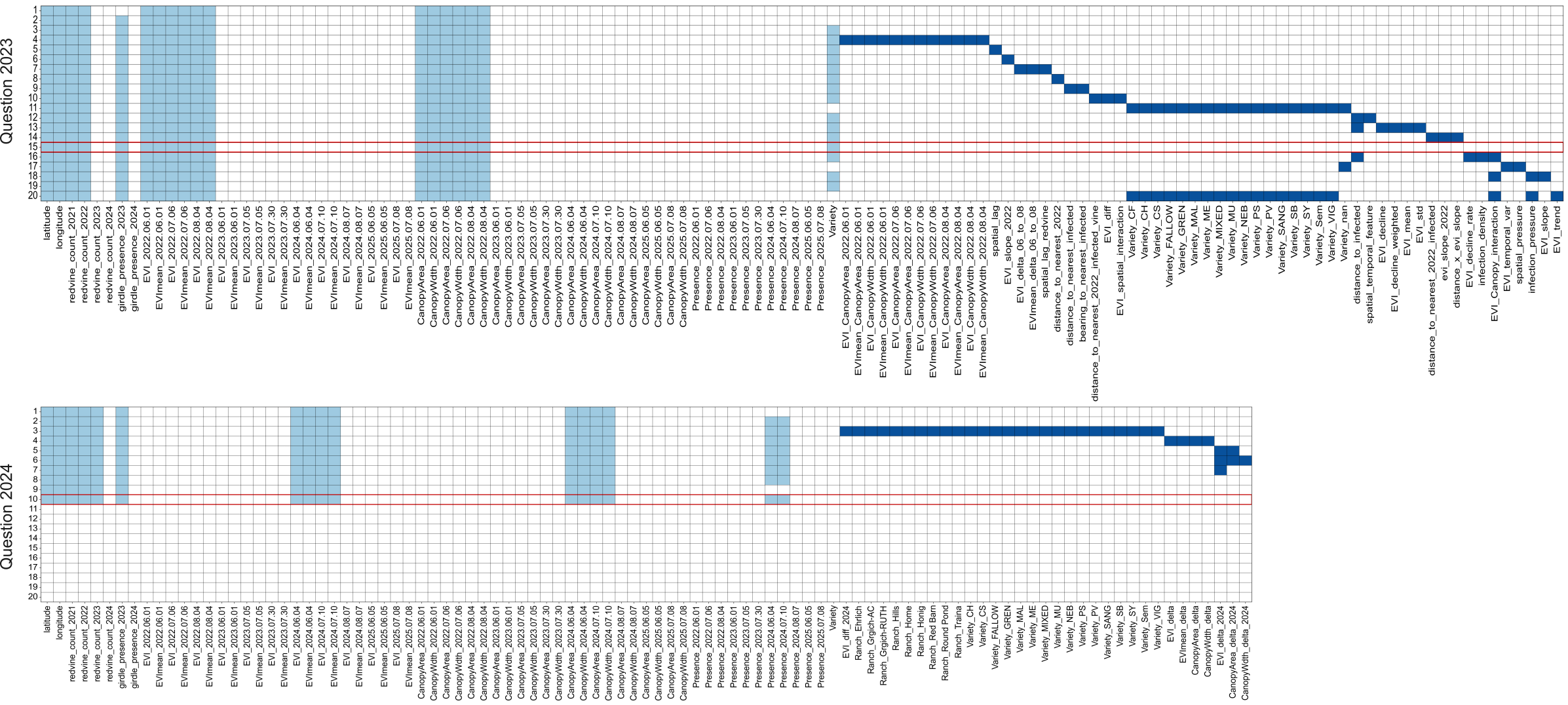}
    \caption{Feature selection maps in Exp2 for the 2023 (top) and 2024 (bottom) prediction questions, respectively. The heatmap illustrates the evolution of feature selection by Aleks across iterations. Light blue indicates raw features from the dataset (ordered consistently across all heatmaps), while dark blue represents derived features, arranged in the order in which they appeared during the experiment.}
    \label{fig4_2}
\end{figure}

The behavioral differences of Aleks between Exp1 and Exp2 highlight the critical role of the DS agent (compare Figure~\ref{fig4_2} with Figure~\ref{fig4_1_1} and Figure~\ref{fig4_1_2}). In the absence of the DS agent, Aleks functioned as a purely data-driven optimizer, prioritizing feature engineering strategies that maximized predictive performance without considering the relevance to plant pathology and disease epidemiology. Under these conditions, Aleks frequently generated new features in the early iterations of an experiment, often relying on statistical correlations with the label rather than on biologically meaningful insights. Consequently, Aleks either maintained a fixed set of original features while continuously proposing new but uninformative ones, as observed in the 2023 prediction task, or terminated prematurely when the DA agent could not identify promising feature directions, as seen in the 2024 prediction task. All these findings suggested the significance role of the DS agent in Aleks to combine both data science and domain-specific knowledge to addressing scientific questions.\\ 

% Model performance of Exp2. Hide them for revisit.
% \begin{table}[h!]
% \centering
% \scriptsize
% \begin{tabular}{l c l c c l c c}
% \hline
% Experiment & Agent Selected Iter. & Model Type & Accuracy & Weighted F1 & Confusion Matrix & R-square & RMSE \\
% \hline
% Exp2 2023     & 15 & Classification & 0.95 & 0.95 & [[3845, 93], [118, 468]] &        &        \\
% Exp2 2024     & 10 & Regression     &      &      &                          & 0.7819 & 0.6128 \\
% \hline
% \end{tabular}
% \caption{Model performance summary.}
% \label{tab:model_performance}
% \end{table}

\subsubsection{Benefits of Full Experiment History as Shared Agent Memory}
\begin{figure}[ht!]
    \centering % Centers the image and caption
    \includegraphics[width=0.5\textwidth]{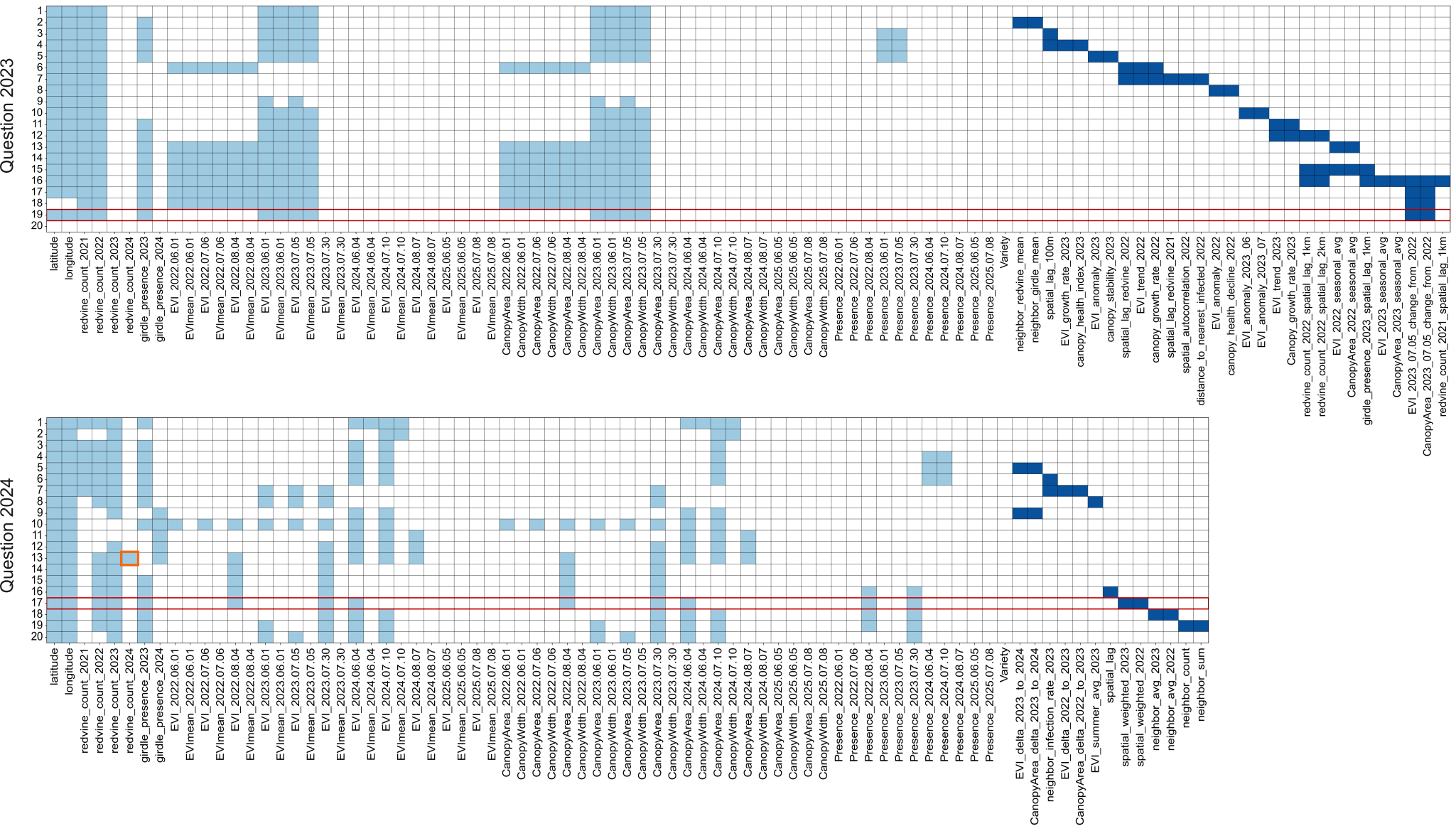}
    \caption{Feature selection maps in Exp3 for the 2023 (top) and 2024 (bottom) prediction questions, respectively. The heatmap illustrates the evolution of feature selection by Aleks across iterations. Light blue indicates raw features from the dataset (ordered consistently across all heatmaps), while dark blue represents derived features, arranged in the order in which they appeared during the experiment.}
    \label{fig4_heatmap_3}
\end{figure}

When the DA agent was restricted to only the current iteration of experiment records in Exp3, Aleks demonstrated less consistency in feature selection (Figure~\ref{fig4_heatmap_3}). Features that neither improved model performance nor aligned with biological knowledge were sometimes repeatedly chosen across iterations within the same run. This occurred because the DA agent, lacking awareness of prior testing, reintroduced features already evaluated in earlier iterations. With no access to the complete experimental history, feature selection became largely probabilistic, as Aleks could only arrive at an effective set by chance rather than through holistic reasoning based on past outcomes. This limitation was particularly evident for features newly proposed by Aleks, which could be suggested repeatedly even when they offered little value for addressing the research question. Moreover, a case of data leakage was identified at iteration 14 in Exp3 for the 2024 prediction question (highlighted in the yellow rectangle in the bottom chart of Figure~\ref{fig4_heatmap_3}). Although corrected in the following iteration, its occurrence underscored the importance of granting the DA agents full access to the shared experimental history. \\

% Model Performance for Exp3. Revisit in the future.
% \begin{table}[ht]
% \centering
% \scriptsize
% \begin{tabular}{l c l c c l c c}
% \hline
% Experiment & Agent Selected Iter. & Model Type & Accuracy & Weighted F1 & Confusion Matrix & R-square & RMSE \\
% \hline
% Exp3 2023     & 19 & Classification & 0.95 & 0.95 & [[3878, 60], [159, 427]] &        &        \\
% Exp3 2024     & 17 & Regression     &      &      &                          & 0.7908 & 0.6001 \\
% \hline
% \end{tabular}
% \caption{Model performance summary.}
% \label{tab:model_performance}
% \end{table}

\subsubsection{Simplified Model Leaderboard as Global Context}
\begin{figure}[ht!]
    \centering % Centers the image and caption
    \includegraphics[width=0.5\textwidth]{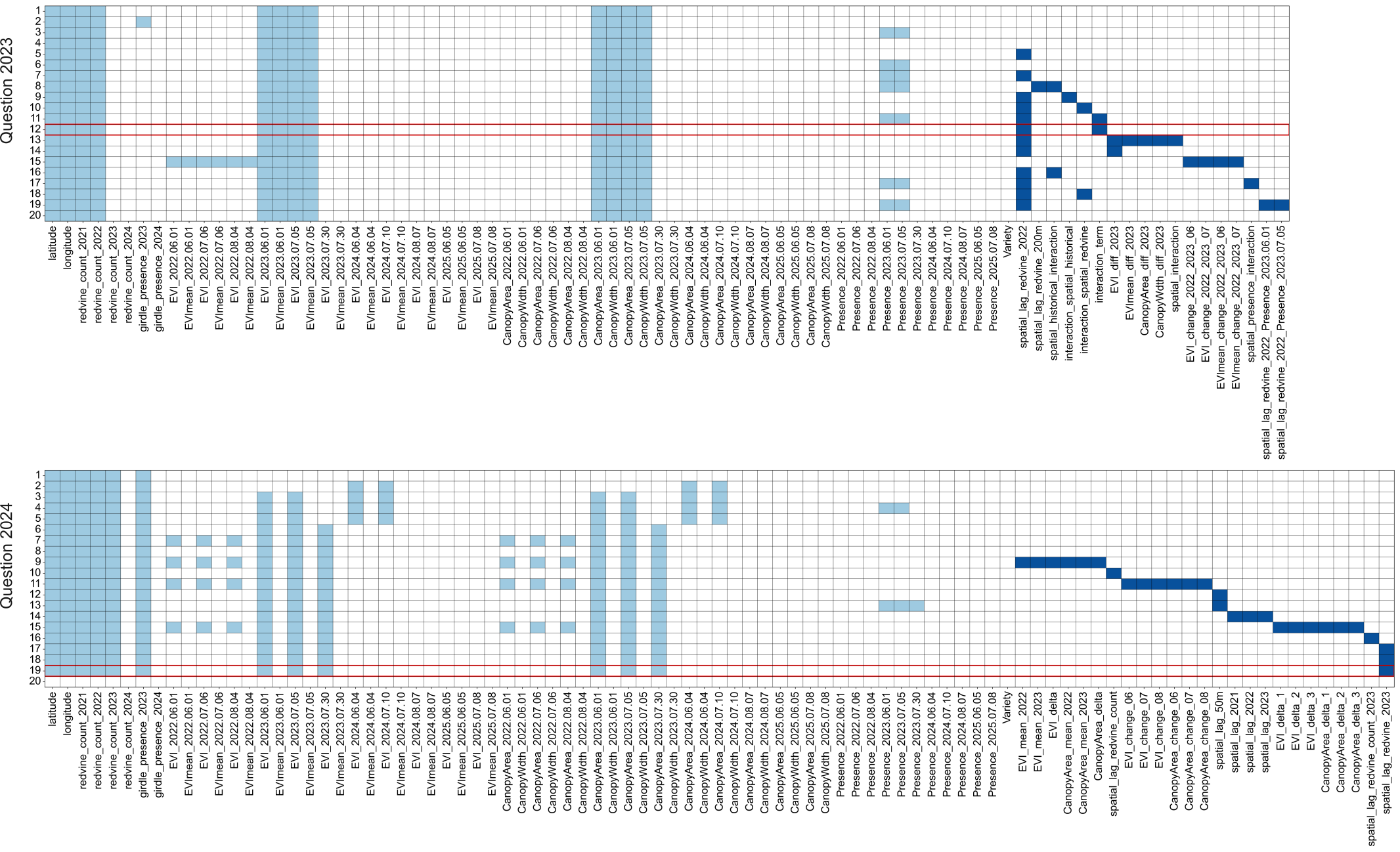}
    \caption{Feature selection maps in Exp4 for the 2023 (top) and 2024 (bottom) prediction questions, respectively. The heatmap illustrates the evolution of feature selection by Aleks across iterations. Light blue indicates raw features from the dataset (ordered consistently across all heatmaps), while dark blue represents derived features, arranged in the order in which they appeared during the experiment.}
    \label{fig4_4}
\end{figure}

In our attempts, adding a simplified model leaderboard to the full experiment history did not produce notable improvements in Aleks’ performance or behavior (Figure~\ref{fig4_4} and Table~\ref{tab:model_performance_exp4}). Both the dynamics of feature selection and the final modeling outcomes were comparable to those observed in Exp1. This result was likely influenced by the specific research questions used in this study. Because all experimental records from 20 iterations could be accommodated within the token limit of the LLM (DeepSeek Chat in this study), the simplified leaderboard did not provide an obvious advantage. In principle, a leaderboard could improve global retrieval of crucial contextual information, but in this case the token limit was not a constraint. Nevertheless, this warrants further investigation, as many plant science problems are far more complex, and their experimental records can easily exceed the token limit of a single LLM. Under such circumstances, maintaining a simplified global context is expected to provide additional benefits for Aleks and similar AI-powered MAS systems.\\

It is worth noting that Aleks’ original report for the 2024 prediction question in Exp4 showed the highest performance among all experimental runs, prompting human scientists to investigate whether any special procedures or feature engineering contributed to this result. Manual verification, however, revealed that the final iteration contained a coding error that conflated performance metrics between the training and testing phases. The second-best model from this run yielded results comparable to others. Although this was the only instance where such a coding bug occurred and went undetected, it highlights an important direction for future research: the development of formal methods to evaluate codebases generated by AI-powered agents.\\

\begin{table}[ht!]
\centering
\scriptsize
\begin{tabular}{l c l c c l c c}
\hline
Experiment & Experiment Iteration Recommended by Aleks & Model Type & R-square & RMSE \\
\hline
Exp4 2023     & 12 & Regression     &      0.7818 & 1.0036 \\
Exp4 2024 (Original)     & 19 & Regression    & 0.9075 & 0.3610 \\
Exp4 2024 (Second Best)    & 18 & Regression     & 0.7584 & 0.4159 \\
\hline
\end{tabular}
\caption{Performance of models determined by Aleks in Exp4 for the 2023 and 2024 prediction questions. For both years, Aleks decided using regression to formulate the question and achieved comparable results with the global model leaderboard.}
\label{tab:model_performance_exp4}
\end{table}

\subsection{Model Generalizability Across Years}
When applying model suggestions from one year to another, the problem formulations and feature engineering strategies of the best models suggested by Aleks demonstrated strong generalizability across years (Table~\ref{tab:model_performance_crossyear}). The best model for the 2023 prediction question was identified in Exp4. This model used a regression formulation and achieved a strong correlation with ground truth counts of GRBD-infected grapevines ($R^2$ = 0.78) with an RMSE of 1. To apply the model to 2024, the codebase was modified only by shifting the corresponding features forward by one year. For example, the feature representing GRBD-infected vine counts in 2022 for the 2023 prediction was replaced with GRBD-infected vine counts in 2023 for the 2024 prediction. All other components, including data preprocessing, feature generation functions, and modeling operations, remained unchanged. Using this adapted codebase, the model achieved a similarly high correlation in 2024 ($R^2$ = 0.79) with an RMSE of 0.65. A comparable trend was observed when applying the best model from the 2024 prediction question to the 2023 dataset. These results confirm that the modeling strategies and feature engineering approaches, guided by both data science and domain-specific knowledge, achieved robust performance across different years. This robustness can be largely attributed to the incorporation of domain knowledge into the modeling process. \\

\begin{table}[ht]
\centering
\scriptsize
\begin{tabular}{l c l c c l c c}
\hline
Modeling Suggestions & Question for Testing & Model Type & R-square & RMSE \\
\hline
2023 (best model across all experiments) & 2024        & Regression & 0.7862 & 0.6451 \\
2024 (best model across all experiments) & 2023    & Regression & 0.7246 & 0.9656 \\
\hline
\end{tabular}
\caption{Model performance for cross year validation. Modeling suggestions include problem formulation, feature engineering options, data preprocessing, and the generated codebase for implementing these suggestions. Questions for testing mean either predicting GRBD infected grapevines in 2023 or 2024. The same dataset and prompt used for other experiments were used for results in this table to avoid potential performance differences due to prompt variation. }
\label{tab:model_performance_crossyear}
\end{table}

\section{Discussion}
\subsection{Advantages of the Aleks}
Aleks, as an AI-powered multi-agent system, offers the advantage of full autonomy in addressing complex scientific questions. Unlike conventional tools that provide narrow outputs, Aleks independently manages the entire reasoning and decision-making process. It not only produces satisfactory final solutions to the questions posed by human scientists but also integrates knowledge across domains such as data science and plant science to reach those solutions. This capacity to reason across multiple layers of expertise allows Aleks to generate scientifically sound outcomes without constant human intervention, thereby expanding the scope and reliability of AI-assisted research. It should be noted that the best models from the case study was used to generate spatial maps to assist human researchers to decide locations for plant tissue sampling in commercial vineyards. However, due to privacy policies, the maps will be shared exclusively upon formal request and growers written consent. \\

Another major advantage of Aleks is its ability to dramatically accelerate the pace of the scientific discovery process. By shortening the experimentation cycle from ideation to implementation and feedback (a couple of hours per full experiment in this study), Aleks increases the throughput of meaningful results. It can rapidly test hypotheses, analyze outcomes, and integrate new feedback into subsequent iterations, which allows for more refined and reliable discoveries within a fraction of the time it would take traditional approaches. This efficiency not only reduces the bottlenecks commonly faced in research workflows by human researchers but also enables scientists to explore a wider range of experimental directions with greater confidence.\\

Beyond technical performance and speed, Aleks opens the door to a paradigm shift in how science can be conducted. Rather than being viewed as a mere tool for analysis, Aleks represents a new mode of scientific discovery that relies on human-AI collaboration. With its capacity to autonomously handle reasoning, experimentation, and integration of cross-domain knowledge, Aleks enables researchers to rethink the structure of scientific practice itself. This integration challenges traditional boundaries between human and AI roles in discovery, ultimately positioning agent-based systems as partners in shaping the future of scientific advancement.\\

\subsection{Limitations of current methods and directions worth investigating}
A major limitation of the current Aleks system arises from the inherent challenges in intelligent agent (IA) and multi-agent system (MAS) design, particularly when built on large language models (LLMs). While LLMs provide powerful generative reasoning capabilities, their limitations in factual accuracy, verification of outputs, and reliability of generated content pose significant risks in the context of scientific discovery. When multiple agents are introduced for cross-verification, achieving consensus among them towards the correct solution is challenging because they may reinforce shared biases and propagate errors or they may argue with each other and cause frequent human intervention. This highlights the need for more rigorous validation mechanisms to ensure that outputs meet the scientific standards required for trustworthy discoveries.\\

Another limitation lies in the constrained tool space currently available to Aleks. At present, its analytical capability is largely bound to auto-sklearn, which restricts flexibility in addressing diverse data formats beyond standard tabular datasets. Scientific research often requires integration of heterogeneous data such as images, time-series, genomic sequences, or spatial maps, as well as access to specialized computational tools developed by human researchers. To address this gap, the system must be expanded to incorporate a broader range of external tools and computational frameworks. Yet, doing so introduces open challenges in ensuring high-quality code generation, robust debugging, and adoption of formal methods for validation to prevent propagation of faulty analyses. Particularly, for data-driven approaches, some coding errors can be hard to find by even human researchers, and such codebase may mislead the result interpretation and ultimate goal to a science question. Besides this, the current Aleks system does not fully integrate hardware components to collect new datasets that can help to address some suggestions raised by the DS agent. It is expected that robotic platforms can be integrated to achieve the full autonomy for experimentation-based research studies.\\

Finally, a critical limitation is the unresolved question of how to structure collaboration between human researchers and AI-powered MAS systems. On one hand, human intervention is necessary to safeguard correctness, ethical considerations, and interpretability. On the other hand, full autonomy is what enables MAS systems like Aleks to deliver high throughput and efficient exploration. Defining the optimal balance between these two extremes is not straightforward and may differ across disciplines, research questions, and even stages within plant science community. Without a clear framework to establish this balance, there is a risk of either over-reliance on the system or excessive human oversight that diminishes the benefits of autonomy.\\

\section{Conclusion}
We introduced Aleks, an AI-powered multi-agent system that autonomously addresses data-driven scientific questions in plant sciences. By integrating domain knowledge, data analysis, and machine learning within a structured framework, Aleks achieved the full autonomy and was able to formulate research questions, design and execute experiments, and iteratively refine solutions without human intervention. In a case study on grapevine red blotch, Aleks identified biologically meaningful features and produced models that balanced predictive performance with interpretability. Ablation studies underscored the importance of domain knowledge and memory in guiding robust and coherent outcomes. Aleks demonstrated its efficacy as an autonomous collaborator for accelerating scientific discovery in plant sciences. Future studies will focus on enhancing the design of multi-agent systems to support trustworthy scientific exploration and on integrating robotic systems to achieve full autonomy across the digital and physical domains of plant science research. We envision that more such initiatives will emerge in the near future, motivating both the plant science community and the broader scientific community to reconsider and reshape the prevailing paradigm of scientific discovery.

\section{Competing interests}
No competing interest is declared. 

\section{Author contributions statement}

Y.J, K.M.G., and D.J. conceived the idea of Aleks, D.J. implemented the Aleks system, D.J., N.G., N.C.J., S.B., K.M.G., and Y.J. proposed the testing study for Aleks, D.J. and N.G. conducted the experiments, D.J., N.C.J., S.B., K.M.G., and Y.J. analyzed the results, D.J., K.M.G., and Y.J. wrote and reviewed the manuscript.  

\section{Acknowledgments}
This study was jointly funded by Cornell AgriTech, Moonshot Grants from the Research and Innovation Office of College of Agriculture and Life Sciences at Cornell, USDA NIFA SCRI, and NASA Acres Consortium.

\bibliographystyle{plain}
\bibliography{reference_aleks}

\begin{thebibliography}{10}

\bibitem{chen2025phenoassistant}
Feng Chen, Ilias Stogiannidis, Andrew Wood, Danilo Bueno, Dominic Williams,
  Fraser Macfarlane, Bruce Grieve, Darren Wells, Jonathan~A Atkinson, Malcolm~J
  Hawkesford, et~al.
\newblock Phenoassistant: A conversational multi-agent ai system for automated
  plant phenotyping.
\newblock {\em arXiv preprint arXiv:2504.19818}, 2025.

\bibitem{cieniewicz2025grapevine}
Elizabeth Cieniewicz and Marc Fuchs.
\newblock Grapevine red blotch disease: A threat to the grape and wine
  industries.
\newblock {\em Annual Review of Virology}, 12, 2025.

\bibitem{cieniewicz2017spatiotemporal}
Elizabeth~J Cieniewicz, Sarah~J Pethybridge, Adrienne Gorny, Laurence~V Madden,
  Heather McLane, Keith~L Perry, and Marc Fuchs.
\newblock Spatiotemporal spread of grapevine red blotch-associated virus in a
  california vineyard.
\newblock {\em Virus Research}, 241:156--162, 2017.

\bibitem{flasco2024investigating}
Madison Flasco, Elizabeth~J Cieniewicz, Monica~L Cooper, Heather McLane, and
  Marc Fuchs.
\newblock Investigating the latency period of grapevine red blotch virus in a
  diseased cabernet franc vineyard experiencing secondary spread.
\newblock {\em American Journal of Enology and Viticulture}, 75(1), 2024.

\bibitem{flasco2023distinct}
Madison~T Flasco, Elizabeth~J Cieniewicz, Sarah~J Pethybridge, and Marc~F
  Fuchs.
\newblock Distinct red blotch disease epidemiological dynamics in two nearby
  vineyards.
\newblock {\em Viruses}, 15(5):1184, 2023.

\bibitem{flasco2023three}
Madison~T Flasco, Victoria Hoyle, Elizabeth~J Cieniewicz, Greg Loeb, Heather
  McLane, Keith Perry, and Marc~F Fuchs.
\newblock The three-cornered alfalfa hopper, spissistilus festinus, is a vector
  of grapevine red blotch virus in vineyards.
\newblock {\em Viruses}, 15(4):927, 2023.

\bibitem{flasco2025decade}
MT~Flasco, DW~Heck, EJ~Cieniewicz, ML~Cooper, SJ~Pethybridge, and MF~Fuchs.
\newblock A decade of grapevine red blotch disease epidemiology reveals zonal
  roguing as novel disease management.
\newblock {\em npj Viruses}, 3(1):29, 2025.

\bibitem{galvan2023scalable}
Fernando E~Romero Galvan, Ryan Pavlick, Graham Trolley, Somil Aggarwal, Daniel
  Sousa, Charles Starr, Elisabeth Forrestel, Stephanie Bolton, Maria del~Mar
  Alsina, Nick Dokoozlian, et~al.
\newblock Scalable early detection of grapevine viral infection with airborne
  imaging spectroscopy.
\newblock {\em Phytopathology{\textregistered}}, 113(8):1439--1446, 2023.

\bibitem{gold2021plant}
Kaitlin~M Gold.
\newblock Plant disease sensing: studying plant-pathogen interactions at scale.
\newblock {\em Msystems}, 6(6):e01228--21, 2021.

\bibitem{jeger2024impact}
Michael Jeger, Robert Beresford, Anna Berlin, Clive Bock, Adrian Fox, Kaitlin~M
  Gold, Adrian~C Newton, Antonio Vicent, and Xiangming Xu.
\newblock Impact of novel methods and research approaches in plant pathology:
  Are individual advances sufficient to meet the wider challenges of disease
  management?
\newblock {\em Plant Pathology}, 73(7):1629--1655, 2024.

\bibitem{jeger2023emerging}
Mike Jeger, Fred Hamelin, and Nik Cunniffe.
\newblock Emerging themes and approaches in plant virus epidemiology.
\newblock {\em Phytopathology{\textregistered}}, 113(9):1630--1646, 2023.

\bibitem{laroche2025grapevine}
E~Laroche-Pinel, K~Singh, M~Flasco, ML~Cooper, M~Fuchs, and L~Brillante.
\newblock Grapevine red blotch virus detection in the vineyard: Leveraging
  machine learning with vis/nir hyperspectral images for asymptomatic and
  symptomatic vines.
\newblock {\em Computers and Electronics in Agriculture}, 234:110251, 2025.

\bibitem{mikaberidze2025cost}
Alexey Mikaberidze, Chaitanya~S Gokhale, Maria Bargu{\'e}s-Ribera, and Prateek
  Verma.
\newblock The cost of fungicide resistance evolution in multi-field plant
  epidemics.
\newblock {\em PLOS Sustainability and Transformation}, 4(6):e0000178, 2025.

\bibitem{rubambiza2023toward}
Gloire Rubambiza, Fernando Romero~Galvan, Ryan Pavlick, Hakim Weatherspoon, and
  Kaitlin~M Gold.
\newblock Toward cloud-native, machine learning base detection of crop disease
  with imaging spectroscopy.
\newblock {\em Journal of Geophysical Research: Biogeosciences},
  128(6):e2022JG007342, 2023.

\bibitem{sawyer2023phenotyping}
Erica Sawyer, Eve Laroche-Pinel, Madison Flasco, Monica~L Cooper, Benjamin
  Corrales, Marc Fuchs, and Luca Brillante.
\newblock Phenotyping grapevine red blotch virus and grapevine
  leafroll-associated viruses before and after symptom expression through
  machine-learning analysis of hyperspectral images.
\newblock {\em Frontiers in plant science}, 14:1117869, 2023.

\bibitem{schmidgall2025agent}
Samuel Schmidgall, Yusheng Su, Ze~Wang, Ximeng Sun, Jialian Wu, Xiaodong Yu,
  Jiang Liu, Michael Moor, Zicheng Liu, and Emad Barsoum.
\newblock Agent laboratory: Using llm agents as research assistants.
\newblock {\em arXiv preprint arXiv:2501.04227}, 2025.

\bibitem{sudarshana2015grapevine}
Mysore~R Sudarshana, Keith~L Perry, and Marc~F Fuchs.
\newblock Grapevine red blotch-associated virus, an emerging threat to the
  grapevine industry.
\newblock {\em Phytopathology}, 105(7):1026--1032, 2015.

\bibitem{swanson2025virtual}
Kyle Swanson, Wesley Wu, Nash~L Bulaong, John~E Pak, and James Zou.
\newblock The virtual lab of ai agents designs new sars-cov-2 nanobodies.
\newblock {\em Nature}, pages 1--3, 2025.

\bibitem{trirat2024automl}
Patara Trirat, Wonyong Jeong, and Sung~Ju Hwang.
\newblock Automl-agent: A multi-agent llm framework for full-pipeline automl.
\newblock {\em arXiv preprint arXiv:2410.02958}, 2024.

\bibitem{yao2023react}
Shunyu Yao, Jeffrey Zhao, Dian Yu, Nan Du, Izhak Shafran, Karthik Narasimhan,
  and Yuan Cao.
\newblock React: Synergizing reasoning and acting in language models.
\newblock {\em International Conference on Learning Representations (ICLR)},
  2023.

\bibitem{zarcoprevisual}
TEJADA Pablo~Jesus ZARCO, Carlos CAMINO, Pieter BECK, Rocio CALDERON, Alberto
  HORNERO, Rocio HERNANDEZ-CLEMENTE, Teja KATTENBORN, Miguel MONTES-BORREGO,
  Leonardo SUSCA, Massimiliano MORELLI, et~al.
\newblock Previsual symptoms of xylella fastidiosa infection revealed in
  spectral plant-trait alterations.
\newblock {\em NATURE PLANTS}.

\bibitem{zarco2021divergent}
Pablo~J Zarco-Tejada, Tom{\'a}s Poblete, Carlos Camino, Victoria
  Gonz{\'a}lez-Dugo, Rocio Calderon, Alberto Hornero, Roc{\'\i}o
  Hern{\'a}ndez-Clemente, Miguel Rom{\'a}n-{\'E}cija, Mar{\'\i}a~Pilar
  Velasco-Amo, Blanca~B Landa, et~al.
\newblock Divergent abiotic spectral pathways unravel pathogen stress signals
  across species.
\newblock {\em Nature Communications}, 12(1):6088, 2021.

\end{thebibliography}

\end{document}